# Shaking Force Balancing of the Orthoglide

Jing Geng[1,2], Vigen Arakelian[1,2], Damien Chablat[1]

[1] LS2N-ECN UMR 6004, 1 rue de la Noë, BP 92101,
F-44321 Nantes, France
[2] INSA Rennes / Mecaproce
20 av. des Buttes de Coesmes, CS 70839,
F-35708 Rennes, France

`jing.geng@insa-rennes.fr`

**Abstract.** The shaking force balancing is a well-known problem in the design of high-speed robotic systems because the variable dynamic loads cause noises, wear and fatigue of mechanical structures. Different solutions, for full or partial shaking force balancing, via internal mass redistribution or by adding auxiliary links were developed. The paper deals with the shaking force balancing of the Orthoglide. The suggested solution based on the optimal acceleration control of the manipulator's common center of mass allows a significant reduction of the shaking force. Compared with the balancing method via adding counterweights or auxiliary substructures, the proposed method can avoid some drawbacks: the increase of the total mass, the overall size and the complexity of the mechanism, which become especially challenging for special parallel manipulators. Using the proposed motion control method, the maximal value of the total mass center acceleration is reduced, as a consequence, the shaking force of the manipulator decreases. The efficiency of the suggested method via numerical simulations carried out with ADAMS is demonstrated.

## 1 Introduction

It is known that a mechanical system with unbalance shaking force/moment transmits substantial vibration to the frame. Thus, a primary objective of the balancing is to cancel or reduce the variable dynamic loads transmitted to the frame and surrounding structures.

The methods of shaking force balancing can be arranged as follows: *i*) by adding counterweight in order to keep the total mass center of moving links stationary [1-2]. It is obvious that the adding of the counterweights is not desirable because it leads to the increase of the total mass, of the overall size and of the efforts in joints; *ii*) by adding auxiliary structures. In [3-5], the parallelograms were used as auxiliary structures in order to create the balanced manipulators. In [6], the pantograph has been added in order to balance the shaking force of Delta robot; *iii*) by installing elastic components [7], [**Erreur ! Source du renvoi introuvable.**]; *iv*) by adjustment of kinematic parameters [**Erreur ! Source du renvoi introuvable.**], [**Erreur ! Source du renvoi introuvable.**]; *v*) via center of mass acceleration control [**Erreur ! Source du renvoi introuvable.**-**Erreur ! Source du renvoi introuvable.**].

The paper deals with the shaking force balancing of the Orthoglie [**Erreur ! Source du renvoi introuvable.**], [**Erreur ! Source du renvoi introuvable.**]. The Orthoglide is a three-degrees-of-freedom parallel manipulator with regular workspace and good compactness. Its three actuators are arranged according to the Cartesian coordinate space (Fig. 1). The moving platform is connected with actuators via three identical kinematical chains. In the paper, it is proposed to apply the shaking balancing via the center of mass acceleration control.

## 2  Shaking force Balancing of the Orthoglide

### 2.1  Problem Formulation

Let us consider the kinematic architecture of the Orthoglide (Fig. 1). It consists of three identical kinematic chains that are formally described as $\underline{P}RP_aR$, where $\underline{P}$, $R$ and $P_a$ denote the actuated prismatic, revolute, and parallelogram joints respectively. The mechanism input is made up by three actuated orthogonal prismatic joints. The output body is connected to the prismatic joint through a set of three kinematic chains. Inside each chain, one parallelogram is used and oriented in a manner that the output body is restricted to translational movements only. The three parallelograms have the same lengths $L = B_iC_i$. The arrangement of the joints in the $\underline{P}RP_aR$ chains has been defined to eliminate any constraint singularity in the Cartesian workspace. Each frame point $A_i$ is fixed on the $i^{th}$ linear axis so that $A_1A_2 = A_1A_3 = A_2A_3$. The points $B_i$ and $C_i$ are located on the $i^{th}$ parallelogram, as is shown in Fig. 1. The reference frame is located at the intersection of the prismatic joint axes and aligns the coordinate axis with them. The details of the design of the Orthoglide and its optimization can be found in [16-18].

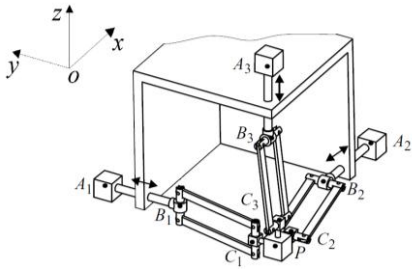
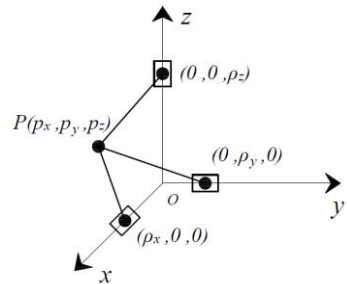

**Fig. 1.** The Orthoglide.   **Fig. 2.** Orthoglide's geometrical model.

For the Orthoglide geometrical model (see Fig. 2), the inverse kinematic equations [18] can be drives in a straightforward way as:

$$\rho_x = p_x + s_x\sqrt{L^2 - p_y^2 - p_z^2} \tag{1}$$

$$\rho_y = p_y + s_y\sqrt{L^2 - p_x^2 - p_z^2} \tag{2}$$

$$\rho_z = p_z + s_z\sqrt{L^2 - p_x^2 - p_y^2} \tag{3}$$

where $s_x$, $s_y$, $s_z$ are the configuration indices that are equal to $\pm 1$; The input vector of the three prismatic joints variables as $\rho = (\rho_x, \rho_y, \rho_z)$ and the output position vector of the tool center point as $p = (p_x, p_y, p_z)$. For the Orthoglide robot, a single inverse kinematic solution is reachable. The shaking forces $\boldsymbol{F}^{sh}$ of mechanisms can be written in the form:

$$\boldsymbol{F}^{sh} = m\ddot{\boldsymbol{s}} \tag{4}$$

where $m = \sum_{i=1}^{n} m_i$ is the total mass of the moving links of the manipulator and $\ddot{\boldsymbol{s}}$ is the acceleration of the total mass centre. As mentioned above, the shaking force balancing via mass redistribution consists in adding counterweights [19] in order to keep the total mass centre of moving links stationary. In this case, $\ddot{\boldsymbol{s}} = 0$ for any configuration of the manipulator and, as a result, the shaking force is cancelled. It is obvious that the adding of supplementary masses as counterweights is not desirable because it leads to the increase of the total mass, of the overall size of the manipulator, the efforts in joints, the shaking moment and the input torques. Therefore, in the present study, it is proposed to minimize the shaking force via reduction of the total mass centre acceleration:

$$\max|\ddot{\boldsymbol{s}}| \to \min_{s(t)} \tag{5}$$

i.e. to apply an optimal control of the total mass centre of moving links that allows one to reduce the maximal value of its acceleration.

For this purpose, let's consider the control of the spatial parallel manipulator Orthoglide through of its centre of mass. To ensure it, let us assume that the centre of mass moves along a straight line between its initial and final positions. Thus, the motion profile used on this path will define the values of shaking forces. For the same displacement of the total centre of mass $S$ and the displacement time $t$, the maximal value of the acceleration changes following the e motion profile [20]: For quantic polynomial profile, the $|a_{max}| = 10S/\sqrt{3}t^2$; For bang-bang profile, $|a_{max}| = 4S/t^2$. It means the application of bang-bang law theoretically brings about a reduction of 31% of the maximal value of the acceleration. Hence, to minimize the maximum value of the acceleration of the total mass centre and, as a result, shaking forces, the "bang-bang" profile should be used. Thus, by reducing the acceleration of the centre of mass of the Orthoglide, a decrease in its shaking forces is achieved. Thus, to achieve the shaking force balancing through the above described approach, it is necessary to consider the relationship between the input parameters and the centre of mass positions of the Orthoglide.

## 2.2 The relationship between the total centre of mass and the input parameters

In order to control the manipulator according to the method described above, it is necessary to establish the relationship between the displacement of the total centre of mass and the input parameters $\boldsymbol{\rho} = (\rho_x, \rho_y, \rho_z)$, i.e. for the given position and the law of motion of the Common Centre of Mass (COM) of the manipulator determine its input displacements. Then, by means of the obtained input parameters via forward kinematics determine the position of the output axis $\mathbf{P}(p_x, p_y, p_z)$. For this purpose, it is necessary to establish the relationship between the common center of mass of the manipulator and its input parameters.

Let us start this issue with the initial and final positions $\mathbf{P}(p_x, p_y, p_z)$ of the platform $\boldsymbol{P_i}(x_i, y_i, z_i)$ and $\boldsymbol{P_f}(x_f, y_f, z_f)$. So, by invers kinematics [18], the input angles corresponding to these positions will be determined: $\boldsymbol{\rho_i}(\rho_{xi}, \rho_{yi}, \rho_{zi})$ and $\boldsymbol{\rho_f}(\rho_{xf}, \rho_{yf}, \rho_{zf})$. The corresponding values of the common COM of the manipulator can also be found: $\mathbf{S_{COM\_i}} = (x_{Si}, y_{Si}, z_{Si})$ and $\mathbf{S_{COM\_f}} = (x_{Sf}, y_{Sf}, z_{Sf})$. The displacement of the total centre of mass is $\mathbf{D}(d_x, d_y, d_z)$. Subsequently, a straight line connecting the initial and final positions of the comment centre mass of the manipulator can be established and its trajectory planning by "bang-bang" profile with the time interval $t_f$ can be ensured: $\mathbf{S_{COM}} = \mathbf{S}(t)$, i.e.

$$\mathbf{S}(t) = \begin{cases} \mathbf{S_{COM\_i}} + 2\left(\dfrac{t}{t_f}\right)^2 \mathbf{D}, (0 \le t \le \dfrac{t_f}{2}) \\ \mathbf{S_{COM\_i}} + \left[-1 + 4\left(\dfrac{t}{t_f}\right) - 2\left(\dfrac{t}{t_f}\right)^2\right] \mathbf{D}, (0 \le t \le \dfrac{t_f}{2}) \end{cases} \quad (6)$$

Let us now consider the relationship between $\mathbf{S_{COM}} = [x(t), y(t), z(t)]$ and the input displacement $\boldsymbol{\rho} = (\rho_x, \rho_y, \rho_z)$.

The common COM of the manipulator can be expressed as:

$$\mathbf{S_{COM}} = \dfrac{\sum_{i=1}^{n} \mathbf{r_i} m_i}{M} \quad (7)$$

where $i$ is the number of the moving link ($i=1,...,n$), $\mathbf{S_{COM}}$ is the coordinate vector of the total mass centre of the manipulator, $\mathbf{r_i}$ is the the coordinate vector of the linkage $i$, $m_i$ is the mass of the linkage $i$.

In the developed prototype, the slider of prismatic joint is designed as body $AB$, where $A$ is not on the three axis but has an offset named $l$. At the same time, $C_1 = C_2 = C_3 = P$. Thus, the coordinates of the joints along X, Y and Z axes are the followings:

X-axis: $C_1 = (p_x, p_y, p_z)$; $B_1 = (\rho_x, 0, 0)$; $A_1 = (\rho_x + l, 0, l)$.

Y-axis: $C_2 = (p_x, p_y, p_z)$; $B_2 = (0, \rho_y, 0)$; $A_2 = (l, \rho_y + l, 0)$.

Z-axis: $C_3 = (p_x, p_y, p_z)$; $B_3 = (0, 0, \rho_z)$; $A_3 = (0, l, \rho_z + l)$.

The mass centers of the parallelograms can be written as: $\left[(x_{C_i} + x_{B_i})/2,\ (y_{C_i} + y_{B_i})/2,\ (z_{C_i} + z_{B_i})/2\right]$, and their mass are $m_1$. The mass center of the three input links are: $\left[(x_{A_i} + x_{B_i})/2,\ (y_{A_i} + y_{B_i})/2,\ (z_{A_i} + z_{B_i})/0\right]$, the masses of input links are denoted as $m_2$. The coordinates of the mass center of the end-effector $P$ are $p_x, p_y, p_z$ and their masses are $m_3$.

With the masses of the corresponding links, the expressions of the total center of mass of the moving links of the Orthoglide can be expressed as:

$$
\begin{aligned}
S_x &= \left[ m_1(\rho_x + 3p_x)/2 + m_2(\rho_x + l) + m_3 p_x \right]/M \\
S_y &= \left[ m_1(\rho_y + 3p_y)/2 + m_2(\rho_y + l) + m_3 p_y \right]/M \\
S_z &= \left[ m_1(\rho_z + 3p_z)/2 + m_2(\rho_z + l) + m_3 p_z \right]/M
\end{aligned}
\qquad (8)
$$

where, $M = 3(m_1 + m_2) + m_3$ is the total mass of the moving links.

According to the proposed method, the displacement of the total center of mass should follow Bang-bang motion profile $\mathbf{S}(t)$, i.e.

$$
x(t) = \begin{cases} x_{Si} + 2\left(\dfrac{t}{t_f}\right)^2 d_x, & (0 \le t \le \dfrac{t_f}{2}) \\ x_{Si} + \left[-1 + 4\left(\dfrac{t}{t_f}\right) - 2\left(\dfrac{t}{t_f}\right)^2\right] d_x, & (0 \le t \le \dfrac{t_f}{2}) \end{cases} \qquad (9)
$$

$$
y(t) = \begin{cases} y_{Si} + 2\left(\dfrac{t}{t_f}\right)()^2 d_y, & (0 \le t \le \dfrac{t_f}{2}) \\ y_{Si} + \left[-1 + 4\left(\dfrac{t}{t_f}\right) - 2\left(\dfrac{t}{t_f}\right)^2\right] d_y, & (0 \le t \le \dfrac{t_f}{2}) \end{cases} \qquad (10)
$$

$$z(t) = \begin{cases} z_{Si} + 2\left(\dfrac{t}{t_f}\right)^2 d_z, (0 \le t \le \dfrac{t_f}{2}) \\ z_{Si} + \left[-1 + 4\left(\dfrac{t}{t_f}\right) - 2\left(\dfrac{t}{t_f}\right)^2\right] d_z, (0 \le t \le \dfrac{t_f}{2}) \end{cases} \quad (11)$$

For each step of the movement, we finally obtain a group of three nonlinear Eq. (12) with 3 unknowns $p_x$, $p_y$ and $p_z$:

$$\begin{cases} (s_x(m_1/2 + m_2)\sqrt{L^2 - p_y^2 - p_z^2} + (2m_1 + m_2 + m_3)p_x + m_2 l)/M = x(t) \\ (s_y(m_1/2 + m_2)\sqrt{L^2 - p_x^2 - p_z^2} + (2m_1 + m_2 + m_3)p_y + m_2 l)/M = y(t) \\ (s_z(m_1/2 + m_2)\sqrt{L^2 - p_x^2 - p_y^2} + (2m_1 + m_2 + m_3)p_z + m_2 l)/M = z(t) \end{cases} \quad (12)$$

Note that the input input displacements $(\rho_x, \rho_y, \rho_z)$ of the manipulator Orthoglide can be found via inverse kinematic Eq. (1), (2) and (3).

## 3     Illustrative example and simulation results

To create a CAD model and carry out simulations via ADAMS software, the following parameters of the Orthoglide are used. These parameters correspond to the parameters of the prototype developed in LS2N (Fig. 6). The geometric parameters are: $L = B_1C_1 = B_2C_2 = B_3C_3 = 0.31m$, $off\_pat = 0.1m$, $s_x = s_y = s_z = 1$. The masses of links are: $m_1 = 0.396 kg$, $m_2 = 0.248 kg$ and $m_3 = 0.905 kg$. The trajectory of the output axis $P$ of the platform is given by its initial position $P_i$ with the coordinates: $x_i = 0$, $y_i = 0$, $z_i = 0$ and the final position $P_f$ with the coordinates: $x_f = -0.1m$, $y_f = 0.07m$, $z_f = -0.11m$. The corresponding input displacements are determined via inverse kinematics: $\rho_{xi} = 0.31m$, $\rho_{xf} = 0.1812472m$, $\rho_{yi} = 0.31m$, $\rho_{yf} = 0.3420294m$, $\rho_{zi} = 0.31m$, $\rho_{zf} = 0.1749561m$. The coordinates of the common COM of the manipulator have been found: $x_{Si} = 0.0360334m$, $y_{Si} = 0.0360334m$, $z_{Si} = 0.0360334m$, $x_{Sf} = -0.0440614m$, $y_{Sf} = 0.0853467m$, $z_{Sf} = -0.0513056m$. The traveling time of this trajectory is $t_f = 1s$.

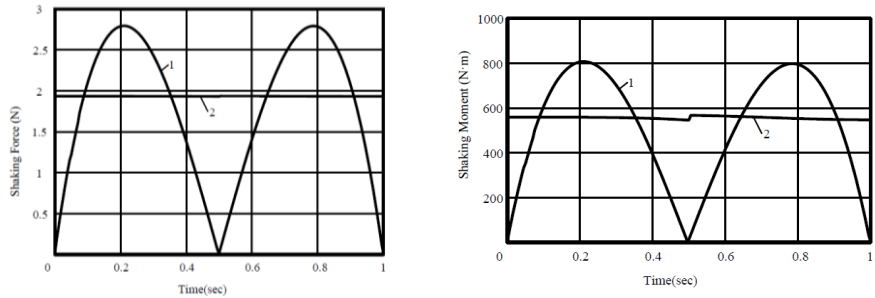

**Fig. 4.** Variations of shaking forces for two studied cases. **Fig. 5.** Variations of shaking moments for two studied cases.

Fig. 3 and Fig. 4 show the variations of shaking forces and the shaking moments for two studied cases: 1) the displacement of the platform of the unbalanced manipulator by the straight line with fifth order polynomial profile and 2) the generation of the motion via the displacement of the manipulator center mass by "bang-bang" profile.

The simulation results show that the shaking force has been reduced up to 31%. Compared to the increase of the shaking moment of the balancing by adding counterweights, the shaking moment has a reduction of 30%. Another advantage of this method is its simplicity and versatility. In the case of changing trajectory, it is just necessary to provide the initial and final coordinates of the end-effector, calculate the input parameters according to the proposed method and implemented in the manipulator control system.

## 4     Conclusion and future works

It is known that the shaking force balancing by counterweights mounted on the links is more appropriate for serial and planar parallel manipulators. It is much more difficult for spatial parallel manipulators. Therefore, in this paper, an alternative method based on optimal acceleration control of the common COM is discussed. The object of study is the spatial 3-DOF parallel manipulator known as Orthoglide. The suggested balancing technique consists in the fact that the Orthoglide is controlled not by applying platform trajectories but by planning the displacements of the total mass center of moving links. The trajectories of the total mass center of the manipulator are defined as straight lines and are parameterized with "bang-bang" profile. Such a control approach allows the reduction of the maximum value of the center of mass. The numerical simulations show the efficiency of the proposed method.

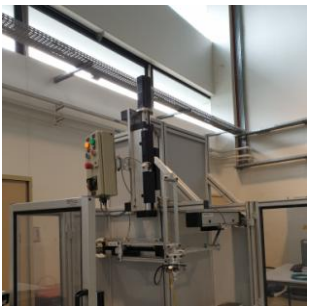

**Fig. 6.** The prototype of the Orthoglide (LS2N).

Future works concern now the experimental validation of the suggested balancing technique via tests that will be carried out on the prototype of the Orthoglide developed in LS2N (Fig. 6).